\documentclass[letterpaper, 10 pt, conference]{ieeeconf}
\IEEEoverridecommandlockouts
\usepackage{cite}
\usepackage{amsmath,amssymb,amsfonts}
\usepackage{algorithmic}
\usepackage{graphicx}
\usepackage{textcomp}
\usepackage{xcolor}
\usepackage{makecell}
\usepackage{mars}
\usepackage[obeyspaces]{url}
\usepackage{hyperref}  

\usepackage{epsfig} 

\usepackage{url}
\usepackage{multirow}
\usepackage{multirow}
\usepackage{threeparttable}
\usepackage{float}
\usepackage{color}
\usepackage{subfigure}

\PassOptionsToPackage{hyphens}{url}

\def\BibTeX{{\rm B\kern-.05em{\sc i\kern-.025em b}\kern-.08em
    T\kern-.1667em\lower.7ex\hbox{E}\kern-.125emX}}

\title{\LARGE \bf The SLAM Hive Benchmarking Suite}
\author{Yuanyuan Yang$^{1}$, Bowen Xu$^{1}$, Yinjie Li$^{1}$ and S\"oren Schwertfeger$^{1}$
\thanks{$^{1}$All authors are with the School of Information Science and Technology, 
	ShanghaiTech University, China.
	{\tt\small [yangyy2, xubw, liyj2@shanghaitech.edu.cn}}%
}
\begin{document}
	
%
%


\marsPublishedIn{Accepted for:} 		

\marsVenue{IEEE Conference on Robotics and Automation (ICRA) 2023}

\marsYear{2023}

\marsPlainAutors{Yuanyuan Yang, Bowen Xu, Yinjie Li and S\"oren Schwertfeger}


\marsMakeCitation{The SLAM Hive Benchmarking Suite}{IEEE Press}


\marsIEEE{}


\makeMARStitle

%
%

\maketitle

\begin{abstract}
	Benchmarking Simultaneous Localization and Mapping (SLAM) algorithms is important to scientists and users of robotic systems alike. But through their many configuration options in hardware and software, SLAM systems feature a vast parameter space that scientists up to now were not able to explore. The proposed SLAM Hive Benchmarking Suite is able to analyze SLAM algorithms in 1000's of mapping runs, through its utilization of container technology and deployment in a cluster. This paper presents the architecture and open source implementation of SLAM Hive and compares it to existing efforts on SLAM evaluation. Furthermore, we highlight the function of SLAM Hive by exploring some open source algorithms on public datasets in terms of accuracy. We compare the algorithms against each other and evaluate how parameters effect not only accuracy but also CPU and memory usage. Through this we show that SLAM Hive can become an essential tool for proper comparisons and evaluations of SLAM algorithms and thus drive the scientific development in the research on SLAM. 
\end{abstract}

\section{INTRODUCTION}

Simultaneous Localization and Mapping (SLAM) is an essential capability for many robotic systems. Consequently, there is lots of research work on SLAM, utilizing various types of sensors and algorithms and employing these to various application scenarios \cite{cadena2016past}. In order to scientifically evaluate the performance of SLAM systems, experiments with them must be reproducible, repeatable and properly compared to other solutions. Running SLAM algorithms with pre-recorded datasets is essential for this, as it allows to repeatedly perform mapping runs with the exact same data. Using ground-truth path information for said datasets it is possible to quantitatively evaluate the localization error of the mapping run, which is generally considered a sufficient measure for the performance of SLAM algorithms \cite{fornasier2021vinseval}.

SLAM is a complex topic and the performance of an approach depends on various factors, including the type of environment, the path the robot took in that environment, sensor types, sensor placement on the robot, settings for the sensor data like frame rate, resolution or maximum range, as well as various configuration parameters for the algorithm like number of particles, number of features or various other thresholds and settings. Furthermore, methods should not only be evaluated against the Absolute Trajectory Error (ATE) and Relative Pose Error (RPE) w.r.t. the ground truth path, but memory consumption, processing time or map quality can also be important factors to consider. 

A robotics engineer who wants to deploy a SLAM system for a specific application should be able to select the best SLAM algorithm, given the number and quality of sensors and computing resources his scenario allows for. Likewise, scientists developing SLAM software should be able to evaluate the performance of their solution under various aspects and configurations. But the number of mapping runs needed to  exhaustively test an algorithm under many permutations of configurations is very big - e.g. comparing 10 algorithms with 5 different configuration each, 10 sensor combinations (e.g. front camera or back camera or both, etc.), 4 different resolutions, 4 different frame rates and 10 different datasets with different environments and amounts of loops, requires 80,000 mapping runs. So currently papers on SLAM only test with and against very few mapping runs.


\begin{figure}[t]
	\centering
	\includegraphics[width=3.5in]{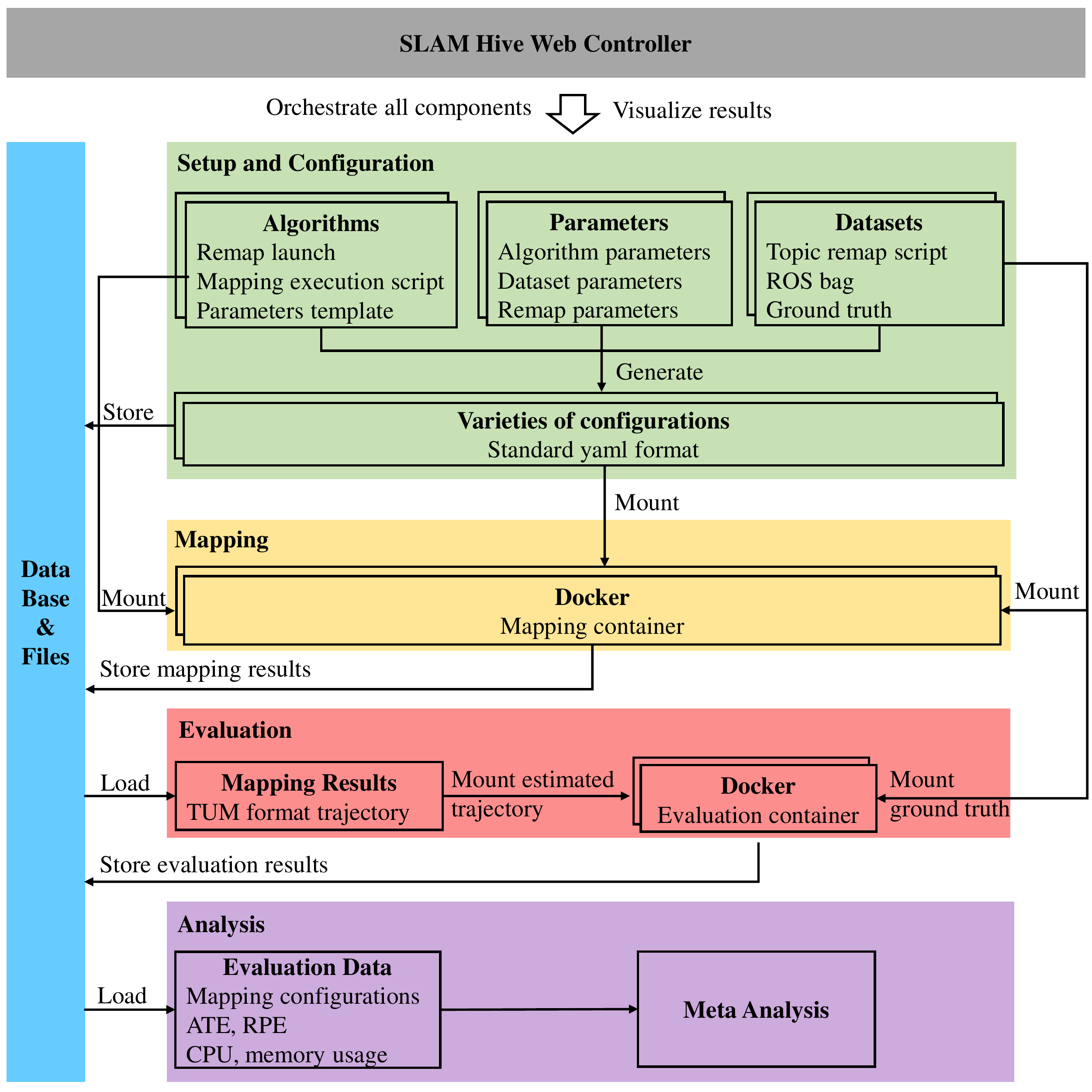}
	\caption{SLAM Hive Benchmarking Suite Overview}
	\label{fig: workflow}
\end{figure}

Thus we propose the SLAM Hive Benchmarking Suite. This suite makes use of Docker containers to run algorithms, evaluation and the whole system itself. It is intended to be deployed in a cluster in order to be able to perform and evaluate 1000's of mapping runs, but can also be used stand-alone on a workstation. Fig. \ref{fig: workflow} shows an overview of the functionality of the system. The suite is still under development, but already ready for evaluating algorithms. Missing components are: the automated creation of many mapping run configurations, the Kubernetes orchestration for deployment in a cluster and the Meta Analysis for analyzing the results of the 1000's of mapping run evaluations. We invite interested researcher to contribute to the development of SLAM Hive, which is available on GitHub \footnote{\url{https://github.com/SLAM-Hive}}.

The key contributions of the ({\it to be completed}) SLAM Hive Benchmarking Suite are:
\begin{itemize}
	\item Defining and implementing a framework performing mapping runs of various SLAM Algorithms in Docker containers, using various configurations and datasets.
	\item Implementing a framework for evaluation of the results of these mapping runs using ({\it different}) evaluation algorithms implemented in Docker containers.
	\item Providing a Web interface, implemented as a Docker container, to view, configure and run the SLAM Hive system and mapping runs.
	\item {\it Configure and run thousands of such mapping runs with different permutations of configurations automatically in a cluster.}
	\item {\it Perform meta analysis of these mapping runs to evaluate them under various aspects (e.g. influence of frame rate, resolution, sensor selection, algorithm parameters, etc.).  }
\end{itemize}

This paper presents the related work in Section \ref{sec:related} and introduces the SLAM Hive Benchmarking Suite in Section \ref{sec:hive} and Section \ref{sec:components}. Its performance is shown with some experiments in Section \ref{sec:experiments} and Section \ref{sec:conclusions} concludes this paper.


\section{Related work}
\label{sec:related}

Datasets are essential for evaluating SLAM systems. There are many well-known public datasets, such as the KITTI dataset \cite{Geiger2013IJRR}, which is widely used in the field of autonomous driving; EuRoC MAV dataset collected by UAV \cite{burri2016euroc}; TUM RGB-D with depth values \cite{sturm2012benchmark}, etc. Among the recently published datasets, \cite{shi2020we} published a service robot dataset that contains some dynamic and daily changing environments scenarios. \cite{wang2020tartanair} presents a challenging data: TartanAir, which is collected in a realistic simulated environment with moving objects, changing light and various weather conditions.

An ideal dataset for SLAM Hive should have many sensors of many different types and pointing in various directions. It should be very high resolution (so it can be down-scaled to simulate low resolution sensors), very high frame rate (which can also simulate low frame-rate systems), hardware synchronized, well calibrated and provide ground truth path data. In \cite{chen2020advanced}  a survey on available robotics-related datasets was conducted, listing the properties of various datasets. It reveals that so far no comprehensive SLAM dataset is available. In \cite{chen2020advanced} we made first steps towards creating such a dataset, which will be concluded with our ShanghaiTech Mapping Robot II soon. This robot, which is already outlined in our paper "Cluster on Wheels" \cite{yang2022cluster}, and the datasets it will create will be integral to fully utilize the capabilities of SLAM Hive.  


There are already some SLAM evaluation systems, for which we show an overview in Table \ref{table: benchmark list}. The famous  KITTI Vision Benchmark Suite \cite{Geiger2012CVPR} provides a set of tools that can evaluate the accuracy of the stereo, optical flow, odometry and object recognition, but it can only use the KITTI dataset of the autonomous driving scenarios they provide and only evaluates the path provided by the user, so it doesn't run the SLAM algorithm itself.

\begin{table*}[t]
	\caption{SLAM Benchmarking systems comparison}
	\label{table: benchmark list}
	\begin{center}
		\begin{threeparttable}
			\begin{tabular}{l|l|l|c|c|c|c|c}
				\hline
				System & Dataset Support & Data Types &  \makecell{Run any alg. \\ in Docker} & \makecell{Monitor\\CPU \& \\Memory} &  \makecell{Automated \\Evaluation} & \makecell{Meta\\Analysis}  & Year \\ 
				\hline
				KITTI Vision \cite{Geiger2012CVPR} & KITTI dataset\cite{Geiger2013IJRR} & mono, stereo, IMU, LiDAR  & &&   & & 2012\\
				TUM RGB-D \cite{sturm2012benchmark} & TUM RGB-D dataset&  RGB-D, IMU & && & & 2012 \\
				EVO \cite{grupp2017evo} & \makecell[l]{Trajectories meet\\ specific formats\tnote{*}} & \makecell[l]{mono, stereo, RGB-D, IMU,\\ LiDAR, etc. } & & && & 2017\\
				SLAMBench \cite{nardi2015introducing,bodin2018slambench2,bujanca2019slambench}& 9 public datasets & mono, stereo, RGB-D, IMU & &  \checkmark & \checkmark && 2019\\
				VINSEval \cite{fornasier2021vinseval}  & Simulation data & mono, stereo, RGB-D, IMU &  & & \checkmark & \checkmark &  2021\\ \hline
				\textbf{SLAM Hive} & any ROS bag & \makecell[l]{mono, stereo, RGB-D, IMU,\\ LiDAR, etc. }& \checkmark & \checkmark & \checkmark & \checkmark & 2022 \\
				\hline
			\end{tabular}
			\begin{tablenotes}
				\scriptsize
				\item[*] EVO is to evaluate the trajectory output by the algorithm. The supported trajectory formats are: 'TUM' trajectory files, 'KITTI' pose files, 'EuRoC MAV', ROS and ROS2 bagfile with geometry\_msgs/PoseStamped, geometry\_msgs/TransformStamped, geometry\_msgs/PoseWithCovarianceStamped or nav\_msgs/Odometry topics or TF messages.
				.
			\end{tablenotes}
		\end{threeparttable}
	\end{center}
	\vspace{-0.6cm}
\end{table*}

In addition to providing the RGB-D dataset, \cite{sturm2012benchmark} also provides automatic evaluation tools both for the evaluation of drift of visual odometry systems and the global pose error of SLAM systems. In that paper, they evaluate the quality of the estimated trajectory instead of the resulting map and accurate ground truth maps. The two well known metrics: RPE and ATE were defined. For both evaluation metrics, they provided easy-to-use evaluation scripts for the users. On this basis, EVO provides executables and a small library for handling, evaluating and comparing the trajectory output of odometry and SLAM algorithms \cite{grupp2017evo}. It supports different trajectory formats and it's easy to output some nice comparison charts.

A series of papers \cite{nardi2015introducing,bodin2018slambench2,bujanca2019slambench} analyzes and compares existing SLAM evaluation systems. SLAMBench proposed that contrast accuracy alone is not enough, it presented a SLAM performance benchmark that combined a framework for quantifying quality-of-result with instrumentation of execution time and energy consumption \cite{nardi2015introducing}. SLAMBench2 proposed that different SLAM applications can have different functional and non-functional requirements. This benchmarking framework evaluates  eight  existing SLAM systems that are lined as a library, over ICL-NUIM, TUM-RGDB and EuRoC MAV datasets, while using a comparable and clearly specified list of performance metrics \cite{bodin2018slambench2} . SLAMBench3 added new support for scene understanding and non-rigid environments (dynamic SLAM), and put forward the corresponding accuracy metrics \cite{bujanca2019slambench} . Although SLAMBench, SLAMBench2 and SLAMBench3 are as comprehensive as possible and support plug and play, it is not as general SLAM Hive w.r.t. adding algorithms, as those need to be linked against SLAMBench, while SLAM Hive runs them in their environment in a container - SLAM Hive could theoretically even support algorithms coded in MatLab and/ or other operating systems.  Moreover, SLAMBench only supports camera (and IMU) data, no LiDAR or other data. 

The TUM VI benchmark published a novel dataset with a diverse set of sequences in different scenes for evaluating visual-inertial odometry (VIO). To verify that the dataset is suitable for benchmarking, it provides the results of three open-source state-of-the-art methods: ROVIO, OKVIS and VINS-Mono, which shows that even the best performing algorithms have significant drift in long (magistrale, outdoors) and visually challenging (slides) sequences \cite{schubert2018tum}. It can be seen that the development of SLAM urgently needs more challenging datasets for SLAM systems evaluation.

Lastly, VINSEval presented a unified framework for statistical relevant evaluation of consistency and robustness of Visual-Inertial Navigation System (VINS) algorithms with fully automated scoreboard generation over a set of selectable attributes \cite{fornasier2021vinseval}. VINSEval uses FlightGoggles, a development environment envisioned to allow the design, implementation, testing and validation of autonomous super vehicles, to generate simulation data with different IMU noisy, amount features, illumination and time delay instead of the real dataset.

According to the studies mentioned above, it can be found that KITTI Vision Bnechmark Suite \cite{Geiger2012CVPR}, RGB-D SLAM systems benchmark \cite{sturm2012benchmark} and the TUM VI benchmark \cite{schubert2018tum} all publish a dataset and provide tools or easy-to-use scripts for evaluating the output of existing SLAM methods. All of them only support using the dataset they provide, which limits the breadth of evaluation of different SLAM algorithms. The SLAMBench series \cite{nardi2015introducing,bodin2018slambench2,bujanca2019slambench} provided more evaluation metrics, supported many algorithms and datasets, but is limited to SLAM algorithms linked against it. Also, it does not support Meta Analysis of many mapping runs, nor performing or creating them automatically. VINSEval \cite{fornasier2021vinseval} provides a robust evaluation method for the visual SLAM community, however it is difficult to ensure the correlation of simulated data with real data and they don't support other sensor data like LiDARs. 


\section{SLAM Hive}
\label{sec:hive}

SLAM Hive is a benchmarking framework that supports mapping execution, evaluation, and performance Meta analysis presentation on webpages with different SLAM methods, datasets and varieties of configurations. All  mapping and evaluation functions are executed in standard Docker containers. For users, it is convenient to reproduce and create some mapping and evaluation tasks. Moreover, many configurations can be customized for comprehensive benchmarks.

\subsection{Overview}
\begin{table*}[t]
	\caption{Supported methods and datasets of SLAM Hive}
	\label{table: support list}
	\vspace{-0.3cm}
	\begin{center}
		\begin{tabular}{l|c|c|c|c|c|c|c}
			\hline
			Method & Monocular & Monocular Inertial & Stereo & Stereo Inertial & RGBD & Lidar Inertial& Year\\ 
			\hline \hline
			ORB-SLAM2 \cite{murORB2}  & \checkmark &  & \checkmark &  & \checkmark &  & 2017\\
			VINS-Mono \cite{8421746}   &  & \checkmark &  &  & & & 2018\\
			VINS-Fusion \cite{DBLP:journals/corr/abs-1901-03638} &   & \checkmark & \checkmark & \checkmark &  &  & 2019\\
			ORB-SLAM3 \cite{9440682} & \checkmark & \checkmark & \checkmark & \checkmark & \checkmark  & & 2021\\
			LIO-SAM \cite{liosam2020shan} &  &  &  & &  & \checkmark & 2020\\
			\hline
		\end{tabular}
	\end{center}
	\begin{center}
		\begin{tabular}{l|c|c|c|c|c|c|l|c}
			\hline
			Dataset & Grayscale camera & RGB camera & RGB-D camera & IMU & GPS & Lidar & Groundtruth & Year\\ 
			\hline \hline
			TUM RGBD \cite{sturm2012benchmark} & & & 1 & \checkmark &  & & Tracking System & 2012\\
			KITTI \cite{Geiger2013IJRR} & 2 & 2 & & \checkmark & \checkmark & \checkmark & Pose via GPS & 2013\\
			EuRoC \cite{burri2016euroc} & & 2 & & \checkmark & & & Tracking System & 2016\\
			\hline
		\end{tabular}
	\end{center}
	\vspace{-0.6cm}
\end{table*}

To the best of our knowledge, SLAM Hive is the first SLAM benchmark framework that has a graphical user interface and automatically presents evaluation results reports. Our system supports custom configurations and some performance metrics to fully evaluate the algorithm. In addition, all task parameters are recorded in the database, which facilitates analysis based on historical tasks and their corresponding parameter values. At the same time, SLAM Hive has a strong scalability, as it can be deployed in a cluster. 

\subsubsection{\textbf{Easy-to-use GUI}}
All operations of SLAM Hive are based on a web GUI. We provide some ready-made Docker images of SLAM methods, generic mapping execution scripts, datasets scripts and some configurations. What users need to do is just select the algorithm and dataset they are interested in, input the desired parameter values, then a new mapping task is created by clicking the button. For systematic evaluations we will support the  automated generation of thousands of permutations of mapping run configurations.

\subsubsection{\textbf{Custom Configurations}}
SLAM Hive already supports many different SLAM methods and datasets that can be seen in Table \ref{table: support list}, and more will be added in the future. These methods could also be closed-source solutions or systems not using ROS, as long as they adhere to the scripting and output standards defined on github. The system supports custom configurations including algorithm parameters, datasets parameters and topic remap.

\subsubsection{\textbf{Comprehensive Evaluation Metrics}}
The accuracy of estimated trajectory is an important metric for evaluating SLAM methods. The evaluation module compares the estimated trajectory from mapping task with the groundtruth, which can plot and compute the trajectory, ATE and RPE. While the CPU and memory usage is monitored during mapping by a monitoring tool called \textit{cAdvisor}. It can perform real-time monitoring and performance data collection of Docker containers, including CPU usage, memory usage etc.

While currently only one evaluation package is supported, we will add more in the future. E.g., we plan to also employ our work that benchmarks the performance of a SLAM system not on the path but on the map its outputs \cite{schwertfeger2016map, hou2022matching}. 


\subsubsection{\textbf{Meta Analysis}}
Since the task information created every time are stored in the database, we can load all historical data from the database, including all parameters value, Min, Max, Mean, Standard Deviation, RMSE of ATE and RPE, CPU and memory usage etc. The system can automatically generate and display various evaluation indicators in the form of tables and pictures. Examples include: "Resolution vs. ATE", "Frame Rate vs. Computation Time", "Number of loops in dataset vs. ATE", "Front camera ATE vs. Back camera ATE", "LiDAR vs. RGB-D". 

\subsubsection{\textbf{Scalability}}
Our system has the standard API, and since all mapping and evaluation tasks are performed in standard Docker containers, whether to expand new algorithms, datasets or even evaluation methods, only the corresponding scripts need to be provided according to the API.

\subsection{Workflow}

SLAM Hive is composed of six modules: web controller, database, configuration generator, mapping container, evaluation container, and meta analysis.

The web controller is responsible for responding to the user's request to create various tasks and presenting a list of algorithms, datasets, parameters, configurations and evaluations. Thousands of configurations can be generated with different algorithms, datasets and parameters by configurations generator.
When a new mapping task is executed, the web controller mounts algorithm execution scripts, parameters and datasets to the corresponding Docker image, and controls the starting and closing of the container. At the same time, the controller records the CPU and memory usage during mapping. After the mapping finishes, the estimated trajectory as well as usage information are saved to database or file system for accuracy evaluation.

Evaluation tasks can also be created by web controller and are executed in standard Docker containers. The evaluated results including estimated trajectory, ATE, RPE are stored in the database after evaluation. Finally, in the future the user will be able to use the meta analysis tool, which loads the required data from the database and aggregates and displays it on the web controller. The workflow of SLAM Hive is shown in the Fig \ref{fig: workflow}.

\section{Main Components of SLAM Hive}
\label{sec:components}

\begin{table*}[t]
	\scriptsize
	\centering
	\tabcolsep=0.14cm
	\renewcommand\arraystretch{1.3}
	\caption{RMSE ATE, CPU and Memory usage of different algorithms under the same sensors combination on EuRoC dataset}
	\label{table: comparison}
	\begin{threeparttable}
		\begin{tabular}{c|l|ccc|ccc|ccc|ccc|ccc}
			\hline
			\multirow{2}*{} & \multirow{2}*{}
			&\multicolumn{3}{c|}{MH\_01\_easy\tnote{1}} & \multicolumn{3}{c|}{MH\_02\_easy\tnote{1}} & \multicolumn{3}{c|}{MH\_03\_medium\tnote{1}} & \multicolumn{3}{c|}{MH\_04\_difficult\tnote{1}} & \multicolumn{3}{c}{MH\_05\_difficult\tnote{1}}\\
			\cline{3-17} && RMSE\tnote{2} & CPU\tnote{3} & RAM\tnote{4} & RMSE & CPU & RAM & RMSE & CPU & RAM & RMSE & CPU & RAM & RMSE & CPU & RAM \\
			
			\hline \hline 
			\multirow{3}*{Monocular}
			&ORB-SLAM2& \textbf{0.043} & 1.36/1.90 & \textbf{978} & 0.038 & 1.38/1.94 & \textbf{954} & \textbf{0.038} & 1.36/1.86 & \textbf{896} & 0.165 & 1.24/1.71 & \textbf{921} & 0.079 & \textbf{1.29/1.79} & \textbf{947}\\
			&ORB-SLAM3& 0.044 & \textbf{1.25/1.67} & 2422 & \textbf{0.037} & \textbf{1.24/1.63} & 2225 & 0.041 & \textbf{1.21/1.53} & 2168 & \textbf{0.054} & \textbf{1.12/1.58} & 1972 & \textbf{0.057} & 1.30/2.51 & 2200\\
			
			\hline \hline 
			\multirow{3}*{\makecell[c]{Monocular\\ Inertial}}
			&VINS-Mono& 0.120 & 1.98/2.56 & 1905 & 0.294 & 1.93/2.57 & 1762 & 0.111 & 1.85/2.50 & 1585 & 0.250 & 1.56/2.40 & 1332 & 0.184 & 1.65/2.50 & \textbf{1477}\\
			&VINS-Fusion& 0.066 & 1.50/2.30 & 2507 & \textbf{0.049} & 1.41/2.23 & \textbf{1375} & \textbf{0.050} & 1.45/2.25 & \textbf{1575} & 0.083 & \textbf{1.26}/2.15 & \textbf{1305} & 0.175 & 1.30/2.19 & 2138\\
			&ORB-SLAM3& \textbf{0.051} & \textbf{1.41}/\textbf{1.85} & \textbf{1746} & 0.065 & \textbf{1.31}/\textbf{1.81} & 2202 & \textbf{0.050} & \textbf{1.20}/\textbf{1.66} & 2364 & \textbf{0.055} & 1.31/\textbf{1.79} & 2036 & \textbf{0.053} & \textbf{1.09}/\textbf{1.55} & 2088\\
			
			\hline \hline 
			\multirow{3}*{Stereo}
			&ORB-SLAM2& 0.038 & 3.24/4.10 & 1387 & \textbf{0.027} & 3.25/3.97 & 1392 & 0.052 & 3.35/4.52 & \textbf{1242} & 0.153 & 3.14/3.92 & \textbf{931} & \textbf{0.067} & 3.14/4.20 & 1599\\
			&VINS-Fusion & 0.053 & \textbf{1.69}/2.60 & \textbf{1265} & 0.054 & \textbf{1.69}/2.46 & \textbf{1393} & 0.082 & \textbf{1.72}/2.79 & 1547 & 0.125 & \textbf{1.52}/2.40 & 1581 & 0.115 & \textbf{1.52}/2.39 & \textbf{1423} \\
			&ORB-SLAM3& \textbf{0.036} & 1.83/\textbf{2.55} & 2731 & 0.034 & 1.77/\textbf{2.45} & 2478 & \textbf{0.044} & 1.90/\textbf{2.62} & 2643 & \textbf{0.062} & 1.59/\textbf{2.37} & 2145 & 0.070 & 1.76/\textbf{3.21} & 2282 \\
			
			\hline \hline 
			\multirow{2}*{\makecell[c]{Stereo\\ Inertial}}
			&VINS-Fusion& 0.043 & \textbf{1.84}/2.66 & \textbf{1535} & \textbf{0.031} & \textbf{1.81}/2.78 & \textbf{2306} & 0.068 & \textbf{1.85}/2.88 & 2833 & 0.093 & \textbf{1.61}/2.55 & \textbf{2045} & 0.090 & \textbf{1.59}/2.59 & \textbf{2038}\\
			&ORB-SLAM3& \textbf{0.041} & 1.91/\textbf{2.41} & 3154 & 0.052 & 1.85/\textbf{2.40} & 2712 & \textbf{0.047} & 1.98/\textbf{2.50} & \textbf{2415} & \textbf{0.046} & 1.98/\textbf{2.45} & 2307 & \textbf{0.054} & 1.99/\textbf{2.46} & 2377 \\
			\hline 
		\end{tabular}
		\begin{tablenotes}
			\scriptsize
			\item[1] They are five sequences of varying difficulty on the EuRoC dataset. \textit{MH\_01\_easy} and \textit{MH\_02\_easy} are with good texture and bright scene, \textit{MH\_03\_medium} is with fast motion and bright scene, \textit{MH\_04\_difficult} and \textit{MH\_05\_difficult} are with fast motion and dark scene.
			\item[2] RMSE is the root mean square error of the ATE, the unit is meter.
			\item[3] The calculation of CPU usage is the sum of all core usage. CPU in the above table is average/maximum CPU usage during mapping, its unit is core. 
			\item[4] RAM is the maximum memory usage during mapping, its unit is MB.
			
		\end{tablenotes}
	\end{threeparttable}
	\vspace{-0.6cm}
\end{table*}

%

\subsection{Visual Web Controller}
The web controller provides the graphical user interface, which is built by the lightweight web application called \textit{Flask}. It is the control center responsible for the creation of mapping and evaluation tasks, as well as the addition of new algorithms, datasets, parameters, configurations. Since the system includes a database, all historical information can be saved. The front end of the system can display the existing data of all modules.

Once a new task is created by the web controller, all of the information including the selection of algorithm and dataset, and configurations with many parameter values are recorded to database. Then a thread can be created at the back end of the controller to start the mapping task and monitor the completion of the task at all times. In the future we will optionally support Kubernetes to deploy the mapping container in a cluster. When the task is completed, a completed notification will be displayed on the front end. In addition, users can add their own new algorithms, datasets and parameters on the web controller by easy step. The evaluation results can be displayed in the form of pictures, tables on the web controller, making it clear to compare the benchmark results.

\subsection{Standard Interface}
Standard interface is an important guarantee that the system has the ability to expand. Our interface consists of standard containers, unified configuration format, algorithm execution scripts, dataset component and standard trajectory format, which are documented on github.

\subsubsection{Standard Containers}
It's difficult to create an unified environment for all algorithms, since different SLAM software require different system environments. Docker is an open source application container engine that can package applications and dependencies into lightweight containers. Using the same Docker image repeatedly will not modify the image itself, ensuring that the same Docker image will be the exact same environment. We provide many SLAM methods and an evaluation Docker image, all mappings and evaluation on SLAM Hive are executed in standard Docker containers. In order to ensure the validity of already performed mapping runs, containers may not be modified after being added to SLAM Hive - if changes are needed one will have to add a new algorithm container. Currently all our algorithms support ROS 1.0, either out of the box or through slight modifications by us. But if needed, the algorithm script (which is included with the algorithm docker) might also extract the data from the ROS datastream to feed it to the algorithm in the format it requires.

\subsubsection{Unified Configuration Format}
When a mapping task is created, a unified configuration is automatically generated in the form of a yaml file, including algorithm selection, dataset selection, algorithm parameters, dataset parameters and ROS remap configuration. The algorithm and dataset execution scripts will extract the required information from the configuration for automated mapping run. 


\subsubsection{Algorithm Execution Scripts}
Algorithm execution scripts includes parameters template and execution script. Before starting the mapping, the execution script gets algorithm parameter values, dataset parameter values and remap information from the configuration file, then generates  custom mapping configuration files according to parameters template and the requirements of the algorithm software, which will be read at the beginning of mapping.

\subsubsection{Dataset component}
The dataset component consists of dataset in the form of ROS bag, groundtruth, and dataset play script. In order to realize custom remap topic, the dataset play script can extract the dataset remap topic from configuration file and generate corresponding dataset play and remap command.

\subsubsection{Standard Trajectory Format}
Most of SLAM algorithms publish the topics information by ROS. So all algorithms containers are built based on ROS in order to replay dataset and record the topic of the estimated pose. For later evaluation, we convert estimated trajectories and groundtruth to the TUM format: $ tx\ ty\ tz\ qx\ qy\ qz\ qw $.

\subsection{Multiple Configurations Mechanism}
Our system has a unified configuration format, varieties of configurations can be easily generated by selecting algorithms, datasets and inputting algorithm parameter values, dataset parameter values and remap topic from scratch or just copying an existing configuration and modifying some of its parameters. These configurations are recorded in database for further performance analysis. The same algorithm and dataset with different algorithm parameters, or the same algorithm and parameters with different datasets may yield interesting results. Nondeterministic algorithms may even be executed several times with the exact same parameters to evaluate their stability.

\section{Benchmarking Experiments}
\label{sec:experiments}

The video accompanying this paper demos how to deploy SLAM Hive on your PC, how to download algorithm containers and datasets with the easy to use scripts, and how to configure, run and analyze a mapping task.

In this section, we used SLAM Hive to complete two sets of benchmark comparison experiments in terms of accuracy, CPU and memory usage. This highlights the capabilities and practicality of SLAM Hive. One is the performance comparison of multiple SLAM methods, the other is the impact of the specific algorithm parameter on the performance of the SLAM method. 

Due to the lack of public datasets rich in various sensors, it is somewhat difficult to compare visual SLAM methods with LiDAR SLAM methods. In the future we will use the ShanghaiTech Mapping Robot II to create such datasets \cite{yang2022cluster}. Therefore, we evaluated the visual SLAM solutions currently supported by the system under different configurations. All tests are on the same device, which is configured as: 16 G memory, Intel Core i7-6700 CPU @ 3.40GHz $\times$ 8.

\subsection{Performance Comparison of SLAM Methods}
\begin{figure}[t]
	\centering
	\subfigure[CPU usage over time]{
		\label{fig:subfig:onefunction} 
		\includegraphics[width=3.4in]{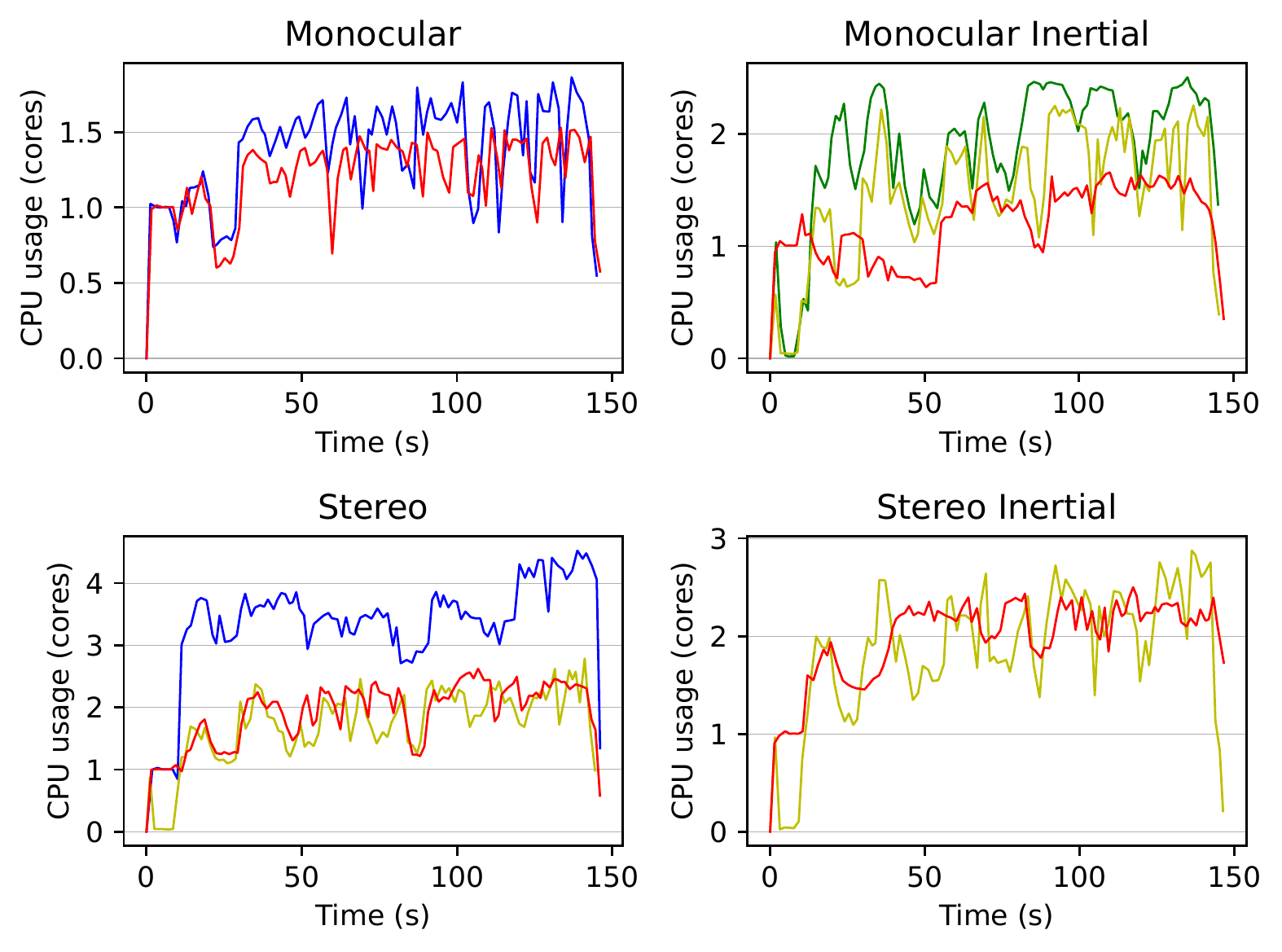}}
	\hspace{0.5in} 
	\subfigure[RAM usage over time]{
		\label{fig:subfig:twofunction} 
		\includegraphics[width=3.4in]{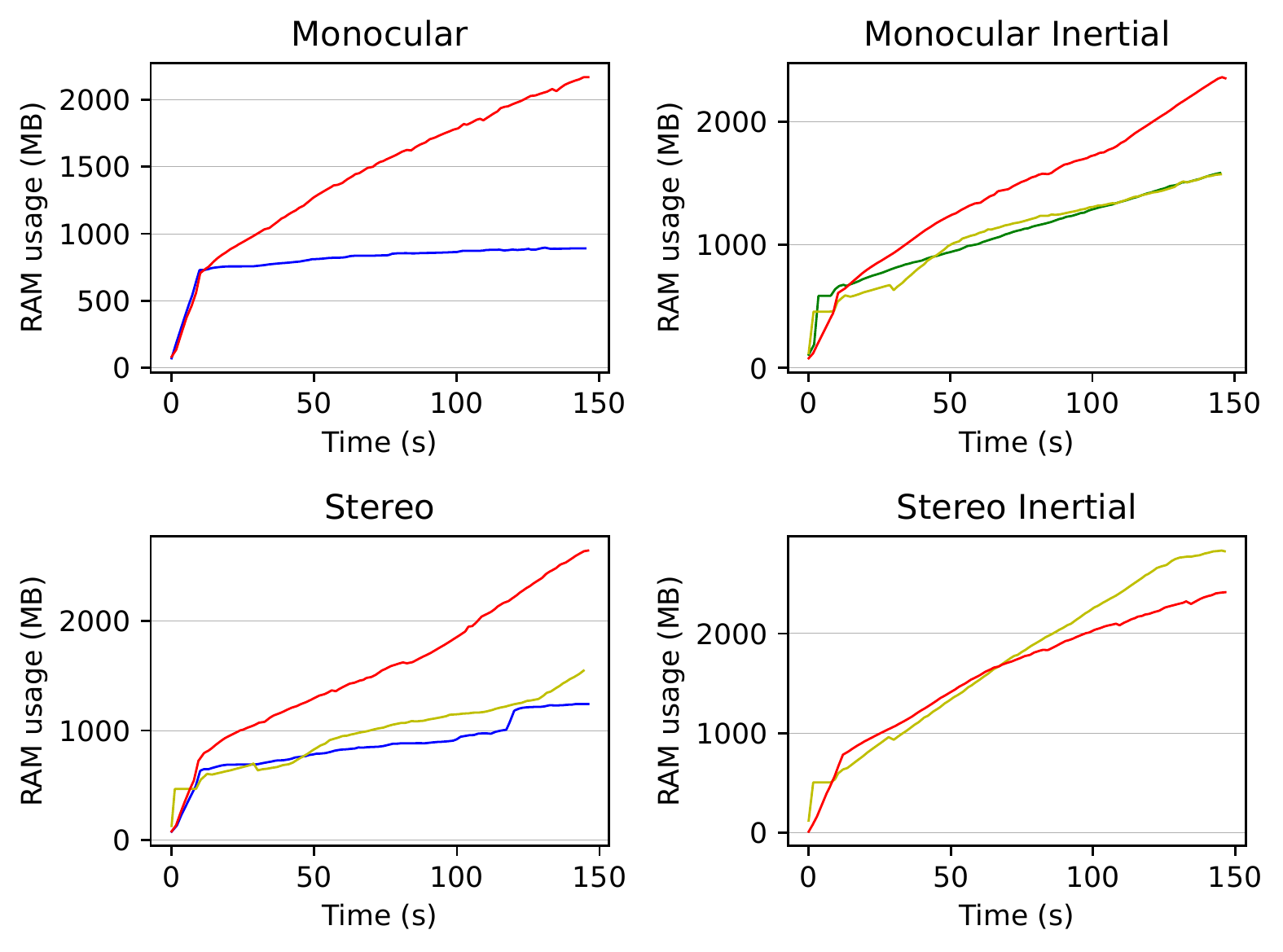}}
	\caption{(a) and (b) are the CPU and memory usage change over time. Both of them are different methods under the same sensors combination on one of sequence on EuRoC dataset: \textit{MH\_03\_medium}. Blue is ORB-SLAM2, red is ORB-SLAM3, green is VINS-Mono, yellow is VINS-Fusion.}
	\label{fig: usage} 
\end{figure}

\begin{figure}[t]
	\centering
	\includegraphics[width=2.9in]{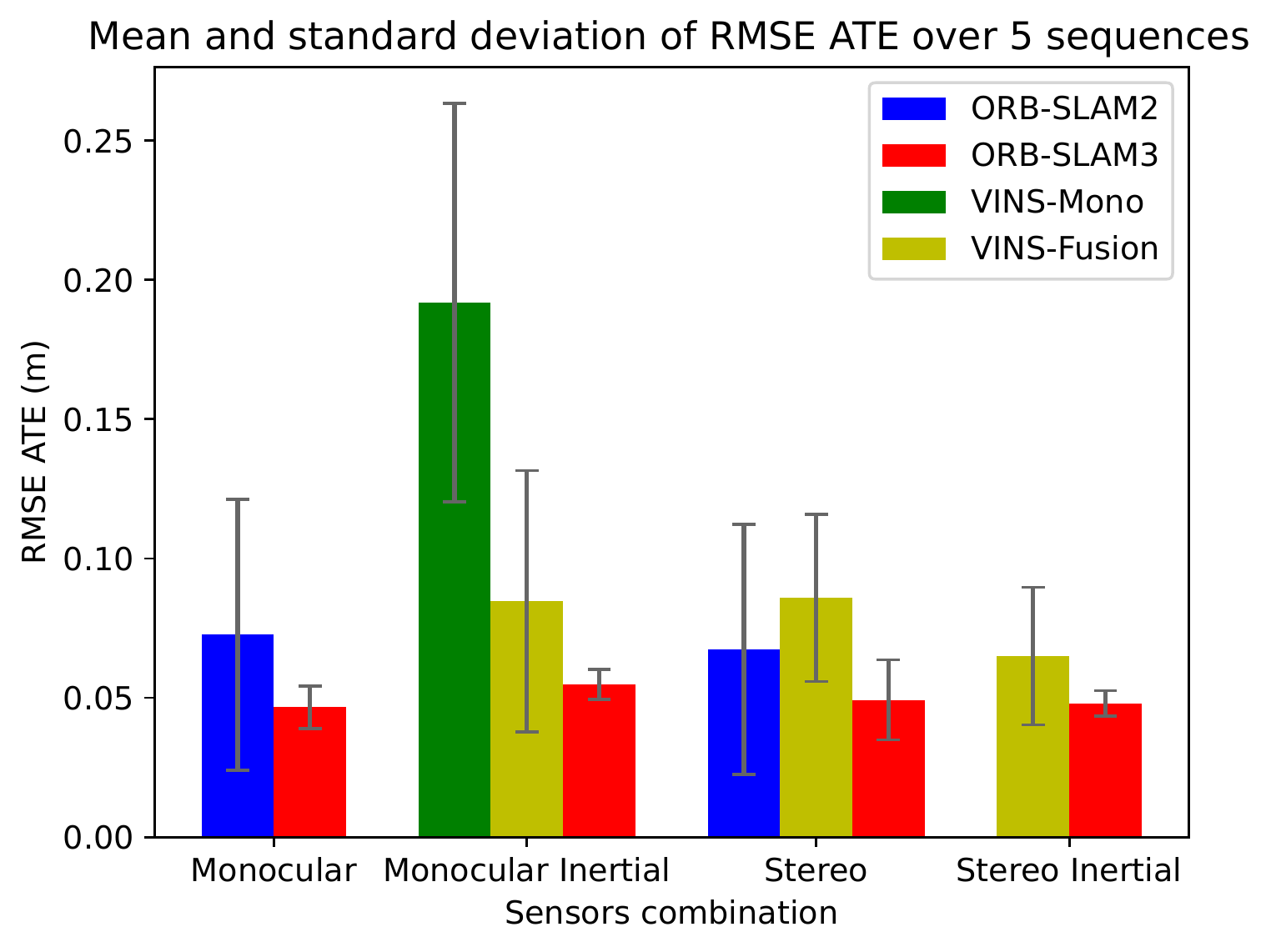}
	\caption{Mean and standard deviation of RMSE ATE over 5 sequences on EuRoC dataset.}
	\label{fig: ate}
\end{figure}

Fig. \ref{fig: ate} shows the comparison of different methods under the same sensor combination on 5 different sequences from the EuRoC dataset. They are increasingly difficult to process in terms of flight dynamics and lighting conditions. For each algorithm, we used the default parameters. The results of each method in each sensor mode are summarized in Table \ref{table: comparison}.

As shown in Fig. \ref{fig: usage}, in the 4 modes, the CPU usage of ORB-SLAM3 is smaller, while the memory usage is larger. In both monocular and stereo modes, the memory usage of ORB-SLAM3 is almost twice that of ORB-SLAM2. Fig. \ref{fig: ate} shows the mean ATE of these methods on 5 sequences in different modes, it is obvious that ORB-SLAM3 performs more stable and has the highest accuracy.

Whether the number of sensor fusion pairs means higher accuracy and more resource consumption is also a topic worth exploring. The CPU usage of ORB-SLAM2 stereo mode is almost twice that of monocular mode. There is no obvious difference in the memory usage of ORB-SLAM3 in the 4 modes, and there is no significant difference in CPU consumption with or without the fusion of the IMU when the number of cameras is the same.

\subsection{Correlation of Parameters to Performance}

Each SLAM method has some parameters of its own and it is difficult to determine how much the change of the parameter value will affect the system. This experiment explores the impact of the number of ORB feature points extracted per image on accuracy, CPU and memory usage. All tests were performed on the \textit{freiburg2/desk} sequence of the TUM-RGBD dataset, a 99.36s, 18.88m, sequence with loop closure.

As shown in the Table \ref{table: test2}, as the number of feature points increases, the ATE and RPE tend to decrease, while the CPU and memory usage tend to increase. When only 250 feature points are extracted, the algorithm fails, indirectly indicating that if the scene has too few features, the performance of ORB-SLAM2 may no longer be robust.

\begin{table}[!ht]
	\vspace{0.3cm}
	\centering
	\caption{ORB-SLAM2 RGB-D comparison under different parameters on freiburg2/desk sequence of TUM-RGBD dataset}
	\label{table: test2}
	\begin{threeparttable}
		\begin{tabular}{c|c|c|c|c}
			\hline
			\makecell[c]{The number of \\ ORB features }& \makecell[c]{RMSE ATE\tnote{1} \\(cm)} & \makecell[c]{RMSE RPE\tnote{1} \\ (cm)} & \makecell[c]{CPU\tnote{1} \\(core)} & \makecell[c]{RAM\tnote{1} \\(MB)}\\ \hline
			250\tnote{2}  & - & - & - & -\\ 
			500  & 0.807 & 1.32 & \textbf{1.26}/\textbf{1.99} & 941 \\ 
			750  & 0.750 & 1.25 & 1.28/2.00 & \textbf{768} \\ 
			1000 & 0.768 & 1.23 & 1.42/2.38 & 880 \\ 
			1250 & 0.693 & 1.16 & 1.64/3.18 & 955 \\ 
			1500 & 0.698 & 1.17 & 1.89/3.37 & 1392 \\ 
			1750 & 0.645 & 1.24 & 2.02/3.42 & 1640 \\ 
			2000 & 0.625 & 1.12 & 2.29/3.75 & 1314 \\ 
			2250 & 0.650 & 1.12 & 2.33/3.87 & 1644 \\ 
			2500 & \textbf{0.623} & \textbf{1.11} & 2.31/3.85 & 1879 \\ \hline
		\end{tabular}
		\begin{tablenotes}
			\scriptsize
			\item[1] Our values are the average of 5 executions. RMSE ATE and RMSE RPE are obtained by aligning with the ground truth trajectory, where RPE is the error of translation per meter.
			\item[2] ORB-SLAM2 RGB-D fails when extracting 250 ORB feature points.
		\end{tablenotes}
	\end{threeparttable}
\end{table}

\section{Conclusions}
\label{sec:conclusions}
The SLAM Hive Benchmarking suite is a containerized system to systematically test and evaluate SLAM algorithms under various configurations using datasets. The system is very scalable and flexible, supporting software using different standards, operating systems and also closed source solutions, as long as scripts can be written to interface with them. The system, for the first time, will be able to explore and analyze the vast space of possible permutations of configurations, datasets and algorithms, when deployed in a cluster. It is also flexible towards the evaluation by implementing these themselves as programs running in their docker container.  Our experiments showed that the suite is already able to properly evaluate SLAM algorithms. 

In the future we plan to complete the feature set presented here. Furthermore, we also plan to explore the real-time performance of algorithms by recording their localization and path estimate on the fly. Systems using GPUs may also be supported, if the cluster is equipped with such. We will also add a user management system and then be able to allow unregistered visitors to use the Meta Analysis function, while registered users with according permissions may add algorithms, datasets and evaluation containers and configure and initiate more mapping runs. 

Through the great flexibility of the SLAM Hive system we may also apply SLAM Hive to similar problems, like benchmarking for extrinsic sensor calibration, or various computer vision tasks like segmentation, object recognition or even prediction. SLAM Hive is available for testing and developing on github and we invite interested programmers to contribute to it.




\section*{ACKNOWLEDGMENT}
%
This work was supported by Science and Technology Commission of Shanghai Municipality (STCSM), project 22JC1410700 \textquotedblleft Evaluation of real-time localization and mapping algorithms for intelligent robots\textquotedblright .

\IEEEtriggeratref{12}
\bibliographystyle{unsrt}
\bibliography{ref}
%

\end{document}